# AND/OR Multi-Valued Decision Diagrams (AOMDDs)
# for Weighted Graphical Models

**Robert Mateescu and Rina Dechter**
{*mateescu,dechter*}@ics.uci.edu
Donald Bren School of Information and Computer Science
University of California, Irvine, CA 92697-3425

## Abstract

Compiling graphical models has recently been under intense investigation, especially for probabilistic modeling and processing. We present here a novel data structure for compiling weighted graphical models (in particular, probabilistic models), called AND/OR Multi-Valued Decision Diagram (AOMDD). This is a generalization of our previous work on constraint networks, to weighted models. The AOMDD is based on the frameworks of AND/OR search spaces for graphical models, and Ordered Binary Decision Diagrams (OBDD). The AOMDD is a canonical representation of a graphical model, and its size and compilation time are bounded exponentially by the treewidth of the graph, rather than pathwidth as is known for OBDDs. We discuss a Variable Elimination schedule for compilation, and present the general APPLY algorithm that combines two weighted AOMDDs, and also present a search based method for compilation method. The preliminary experimental evaluation is quite encouraging, showing the potential of the AOMDD data structure.

## 1  Introduction

We present here an extension of AND/OR Multi-Valued Decision Diagrams (AOMDDs) [13] to general weighted graphical models, including Bayesian networks, influence diagrams and Markov random fields.

The work on AOMDDs is based on two existing frameworks: (1) AND/OR search spaces for graphical models and (2) decision diagrams (DD). AND/OR search spaces [9] have proven to be a unifying framework for various classes of search algorithms for graphical models. The main characteristic is the exploitation of independencies between variables during search, which can provide exponential speedups over traditional search methods that can be viewed as traversing an OR structure. The AND nodes capture problem decomposition into *independent subproblems*, and the OR nodes represent branching according to variable values.

Decision diagrams are widely used in many areas of research, especially in software and hardware verification [5]. A BDD represents a Boolean function by a directed acyclic graph with two sink nodes (labeled 0 and 1), and every internal node is labeled with a variable and has exactly two children: *low* for 0 and *high* for 1. A BDD is *ordered* if variables are encountered in the same order along every path. A BDD is *reduced* if all isomorphic nodes (i.e., with the same label and identical children) are merged, and all redundant nodes (i.e., whose *low* and *high* children are identical) are eliminated. The result is the celebrated *reduced ordered binary decision diagram*, or OBDD [3].

AOMDDs combine the two ideas, in order to create a decision diagram that has an AND/OR structure, thus exploiting problem decomposition. As a detail, the number of values is also increased from two to any constant, but this is less significant for the algorithms.

A decision diagram offers a compilation of a problem. It typically requires an extended offline effort in order to be able to support polynomial (in its size) or constant time online queries. The benefit of moving from OR structure to AND/OR is in a lower complexity of the algorithms and size of the compiled structure. It typically moves from being bounded exponentially in *pathwidth* $pw^*$, which is characteristic to chain decompositions or linear structures, to being exponentially in *treewidth* $w^*$, which is characteristic of tree structures (it always holds that $w^* \leq pw^*$ and $pw^* \leq w^* \cdot \log n$).

Our contributions in this paper are as follows. (1) We formally describe the extension of AND/OR multi-valued decision diagram (AOMDD) to weighted graphical models. (2) We describe the extension to weighted models of the APPLY operator that combines two AOMDDs by an operation. The output of APPLY is still bounded by the product of the sizes of the inputs. (3) We present two compilation algorithms for AOMDDs. One is based on the repeated application of APPLY along a Variable Elimination schedule. The other is based on search. Both schemes are exponential in the treewidth of the model. (4) We provide encouraging preliminary experimental evaluation of the search based compilation method. (5) We discuss how AOMDDs relate to various earlier and recent works, providing a unifying perspective for all these methods.

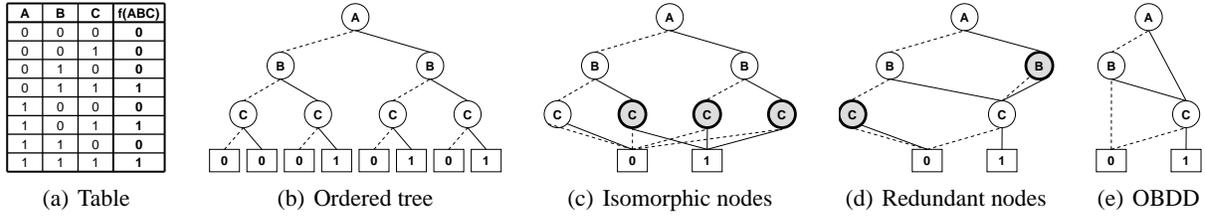

Figure 1: Boolean function representations

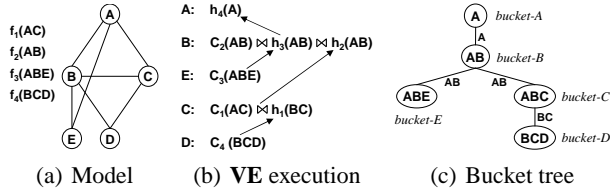

Figure 2: Variable Elimination

The structure of the paper is as follows: Section 2 provides preliminaries. Section 3 gives an overview of AND/OR search space. Section 4 describes the AOMDD for constraint networks, the Variable Elimination schedule for compilation and the APPLY operator, and a search based compilation scheme. Section 5 contains the main contribution: the extension of AOMDDs to weighted models, and a discussion of their canonical form and the extensions of the compilation schedule and APPLY operator. Section 6 provides experimental evaluation and section 7 concludes.

## 2 Preliminaries

In this section we describe graphical models, Binary Decision Diagrams (OBDDs) and Variable Elimination.

DEFINITION 1 (graphical model) *A graphical model $\mathcal{R}$ is a 4-tuple, $\mathcal{R} = \langle \mathbf{X}, \mathbf{D}, \mathbf{F}, \otimes \rangle$, where: (1) $\mathbf{X} = \{X_1, \dots, X_n\}$ is a set of variables; (2) $\mathbf{D} = \{D_1, \dots, D_n\}$ is the set of their respective finite domains of values; (3) $\mathbf{F} = \{f_1, \dots, f_r\}$ is a set of discrete real-valued functions, each defined over a subset of variables $S_i \subseteq \mathbf{X}$, called its scope, and sometimes denoted by $scope(f_i)$. (4) $\otimes_i f_i \in \{\prod_i f_i, \sum_i f_i, \bowtie_i f_i\}$ is a combination operator[1]. The graphical model represents the combination of all its functions: $\otimes_{i=1}^r f_i$. A reasoning task is*

---

[1]The combination operator can be defined axiomatically [17].

*based on a projection (elimination) operator, $\Downarrow$, and is defined by: $\Downarrow_{Z_1} \otimes_{i=1}^r f_i, \dots, \Downarrow_{Z_t} \otimes_{i=1}^r f_i$, where $Z_i \subseteq \mathbf{X}$.*

Examples of graphical models include Bayesian networks, constraint networks, influence diagrams, Markov networks.

DEFINITION 2 (universal equivalent graphical model) *Given a graphical model $\mathcal{R} = \langle \mathbf{X}, \mathbf{D}, \mathbf{F}_1, \otimes \rangle$ the universal equivalent model of $\mathcal{R}$ is $u(\mathcal{R}) = \langle \mathbf{X}, \mathbf{D}, \mathbf{F}_2 = \{\otimes_{f_i \in \mathbf{F}_1} f_i\}, \otimes \rangle$.*

Two graphical models are **equivalent** if they represent the same set of solutions. Namely, if they have the same universal model.

DEFINITION 3 (primal graph) *The primal graph of a graphical model is an undirected graph that has variables as its vertices and an edge connects any two variables that appear in the scope of the same function.*

A pseudo tree resembles the tree rearrangements [11]:

DEFINITION 4 (pseudo tree) *A pseudo tree of a graph $G = (\mathbf{X}, E)$ is a rooted tree $\mathcal{T}$ having the same set of nodes $\mathbf{X}$, such that every arc in $E$ is a back-arc in $\mathcal{T}$ (i.e., it connects nodes on the same path from root).*

DEFINITION 5 (induced graph, induced width, treewidth, pathwidth) *An ordered graph is a pair $(G, d)$, where $G$ is an undirected graph, and $d = (X_1, \dots, X_n)$ is an ordering of the nodes. The width of a node in an ordered graph is the number of neighbors that precede it in the ordering. The width of an ordering $d$, denoted by $w(d)$, is the maximum width over all nodes. The induced width of an ordered graph, $w^*(d)$, is the width of the induced ordered graph obtained as follows: for each node, from last to first in the ordering, its preceding neighbors are connected in a clique. The induced width of a graph, $w^*$, is the minimal induced width over all orderings. The induced width is also equal to the treewidth of a graph. The pathwidth $pw^*$ of a graph is the treewidth over the restricted class of orderings that correspond to chain decompositions.*

### 2.1 Binary Decision Diagrams

Decision diagrams are widely used in many areas of research to represent decision processes. In particular, they can be used to represent functions. Due to the fundamental importance of Boolean functions, a lot of effort has been dedicated to the study of *Binary Decision Diagrams* (BDDs), which are extensively used in software and hardware verification [5, 15, 12, 3].

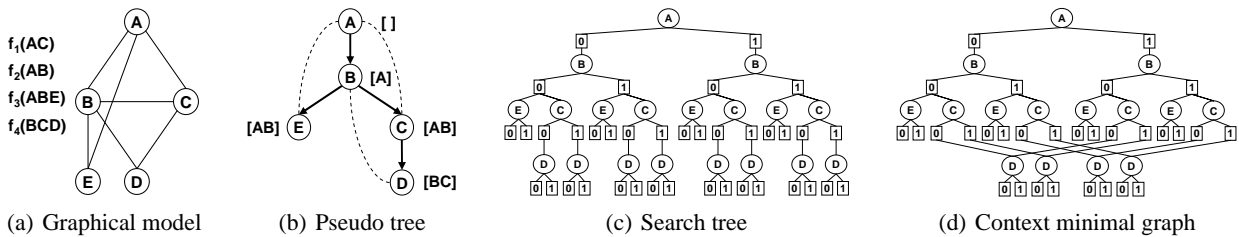

(a) Graphical model      (b) Pseudo tree      (c) Search tree      (d) Context minimal graph

Figure 3: AND/OR search space

A BDD is a representation of a Boolean function. Given $\mathbb{B} = \{0, 1\}$, a Boolean function $f : \mathbb{B}^n \rightarrow \mathbb{B}$, has $n$ arguments, $X_1, \cdots, X_n$, which are Boolean variables, and takes Boolean values. A Boolean function can be represented by a table (see Figure 1(a)), but this is exponential in $n$, and so is the binary tree representation in Figure 1(b). The goal is to have a compact representation, that also supports efficient operations between functions. OBDDs [3] provide such a framework by imposing the same order to the variables along each path in the binary tree, and then applying the following two reduction rules exhaustively:

(1) *isomorphism*: merge nodes that have the same label and the same respective children (see Figure 1(c)).
(2) *redundancy*: eliminate nodes whose low (zero) and high (one) edges point to the same node, and connect the parent of removed node directly to the child of removed node (see Figure 1(d)).

The resulting OBDD is shown in Figure 1(e).

### 2.2 Variable Elimination (VE)

Variable elimination (**VE**) [2, 8] is a well known algorithm for inference in graphical models. Consider a graphical model $\mathcal{R} = \langle \mathbf{X}, \mathbf{D}, \mathbf{F} \rangle$ and an elimination ordering $d = (X_1, X_2, \ldots, X_n)$ ($X_n$ is eliminated first, $X_1$ last). Each function placed in the bucket of its latest variable in $d$. Buckets are processed from $X_n$ to $X_1$ by eliminating the bucket variable (the functions residing in the bucket are combined together, and the bucket variable is projected out) and placing the resulting function (also called *message*) in the bucket of its latest variable in $d$. Figure 2(a) shows a graphical model and 2(b) the execution of **VE**.

**VE** execution defines a bucket tree, by linking the bucket of each $X_i$ to the destination bucket of its message (called the *parent bucket*). A node in the bucket tree has a *bucket variable*, a *collection of functions*, and a *scope* (the union of the scopes of its functions). If the nodes of the bucket tree are replaced by their respective bucket variables, we obtain a pseudo tree (see Figure 2(c) and 3(b)).

## 3 AND/OR Search Space

The AND/OR search space [9] is a recently introduced unifying framework for advanced algorithmic schemes for graphical models. Its main virtue consists in exploiting independencies between variables during search, which can provide exponential speedups over traditional search methods oblivious to problem structure.

### 3.1 AND/OR Search Trees

Given a graphical model $\mathcal{M} = \langle \mathbf{X}, \mathbf{D}, \mathbf{F} \rangle$, its primal graph $G$ and a pseudo tree $\mathcal{T}$ of $G$, the associated AND/OR search tree, $S_{\mathcal{T}}(\mathcal{R})$, has alternating levels of OR and AND nodes. The OR nodes are labeled $X_i$ and correspond to the variables. The AND nodes are labeled $\langle X_i, x_i \rangle$ and correspond to the value assignments in the domains of the variables. The structure of the AND/OR search tree is based on the underlying pseudo tree $\mathcal{T}$. The root of the AND/OR search tree is an OR node labeled with the root of $\mathcal{T}$. The children of an OR node $X_i$ are AND nodes labeled with assignments $\langle X_i, x_i \rangle$ that are consistent with the assignments along the path from the root. The children of an AND node $\langle X_i, x_i \rangle$ are OR nodes labeled with the children of variable $X_i$ in the pseudo tree $\mathcal{T}$.

The AND/OR search tree can be traversed by a depth first search algorithm, thus using linear space. It was already shown [11, 1, 6, 9] that:

**THEOREM 1** *Given a graphical model $\mathcal{M}$ and a pseudo tree $\mathcal{T}$ of depth $m$, the size of the AND/OR search tree based on $\mathcal{T}$ is $O(n\,k^m)$, where $k$ bounds the domains of variables. A graphical model having treewidth $w^*$ has a pseudo tree of depth at most $w^* \log n$, therefore it has an AND/OR search tree of size $O(n\,k^{w^* \log n})$.*

### 3.2 AND/OR Search Graphs

The AND/OR search tree may contain nodes that root identical conditioned subproblems. These nodes are said to be *unifiable*. When unifiable nodes are merged, the search space becomes a graph. Its size becomes smaller at the expense of using additional memory by the search algorithm. The depth first search algorithm can therefore be modified to cache previously computed results, and retrieve them when the same nodes are encountered again. Some unifiable nodes can be identified based on their *contexts*. We can define graph based contexts for both OR nodes and AND nodes, just by expressing the set of ancestor variables in $\mathcal{T}$ that completely determine a conditioned subproblem. However, it can be shown that using caching based on OR contexts makes caching based on AND contexts redundant, so we will only use *OR caching*.

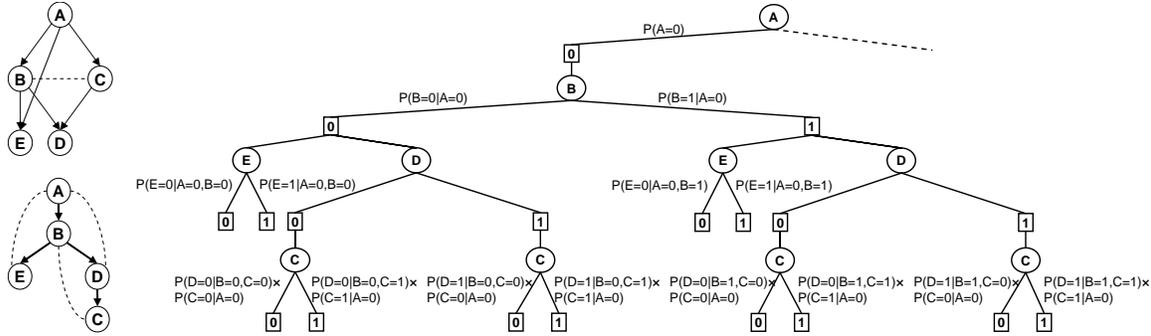

Figure 4: Arc weights for probabilistic networks

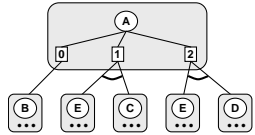

Figure 5: Meta-node

Given a pseudo tree $\mathcal{T}$ of an AND/OR search space, the *context* of an OR node $X$, denoted by $context(X) = [X_1 \dots X_p]$, is the set of ancestors of $X$ in $\mathcal{T}$ ordered descendingly, that are connected in the primal graph to $X$ or to descendants of $X$.

It is easy to verify that the context of $X$ separates the subproblem below $X$ from the rest of the network. The *context minimal* AND/OR graph is obtained by merging all the context unifiable OR nodes. It was shown that [1, 9]:

**THEOREM 2** *Given a graphical model $\mathcal{M}$, its primal graph $G$ and a pseudo tree $\mathcal{T}$, the size of the context minimal AND/OR search graph based on $\mathcal{T}$ is $O(n\,k^{w_{\mathcal{T}}^*(G)})$, where $w_{\mathcal{T}}^*(G)$ is the induced width of $G$ over the depth first traversal of $\mathcal{T}$, and $k$ bounds the domain size.*

Figure 3(a) shows the primal graph of a graphical model defined by the functions $f_1, \dots, f_4$, which are assumed to be strictly positive (i.e., every assignment is valid). Figure 3(b) shows a pseudo tree for the graph. The dotted lines are edges in the primal graph, and back-arcs in the pseudo-tree. The OR context of each node is shown in square brackets. Figure 3(c) shows the AND/OR search tree and 3(d) shows the context minimal AND/OR graph.

### 3.3 Weighted AND/OR Search Graphs

In some cases (e.g. constraint networks), the functions of the graphical model take binary values (0 and 1, or *true* and *false*). In this case, an AND/OR search graph expresses the consistency (valid or not) of each assignment, and can associate this value with its leaves.

In more general cases, which are the focus of this paper, the functions of the graphical model take (positive) real values, called *weights*. For example, in Bayesian networks the weights express the conditional probability. In the more general case of weighted models, it is useful to associate weights to the internal OR-AND arcs in the AND/OR graph, to maintain the global function decomposition and facilitate the merging of nodes.

**DEFINITION 6 (buckets relative to a backbone tree)**
*Given a graphical model $\mathcal{R} = \langle \mathbf{X}, \mathbf{D}, \mathbf{F}, \otimes \rangle$ and a backbone tree $\mathcal{T}$, the bucket of $X_i$ relative to $\mathcal{T}$, denoted by $B_{\mathcal{T}}(X_i)$, is the set of functions whose scopes contain $X_i$ and are included in $path_{\mathcal{T}}(X_i)$, which is the set of variables from the root to $X_i$ in $\mathcal{T}$. Namely, $B_{\mathcal{T}}(X_i) = \{f \in F | X_i \in scope(f), scope(f) \subseteq path_{\mathcal{T}}(X_i)\}$.*

**DEFINITION 7 (OR-AND weights)** *Given an AND/OR tree $S_{\mathcal{T}}(\mathcal{R})$, of a graphical model $\mathcal{R}$, the weight $w_{(n,m)}(X_i, x_i)$ of arc $(n, m)$ where $X_i$ labels $n$ and $x_i$ labels $m$, is the combination (e.g. product) of all the functions in $B_{\mathcal{T}}(X_i)$ assigned by values along the path to $m$, $\pi_m$. Formally, $w_{(n,m)}(X_i, x_i) = \otimes_{f \in B_{\mathcal{T}}(X_i)} f(asgn(\pi_m)[scope(f)])$.*

Figure 4 shows a belief network, a DFS tree that drives its weighted AND/OR search tree, and a portion of the AND/OR search tree with the appropriate weights on the arcs expressed symbolically. In this case the bucket of $E$ contains the function $P(E|A, B)$, and the bucket of $C$ contains two functions, $P(C|A)$ and $P(D|B, C)$. Note that $P(D|B, C)$ belongs neither to the bucket of $B$ nor to the bucket of $D$, but it is contained in the bucket of $C$, which is the last variable in its scope to be instantiated in a path from the root of the pseudo tree.

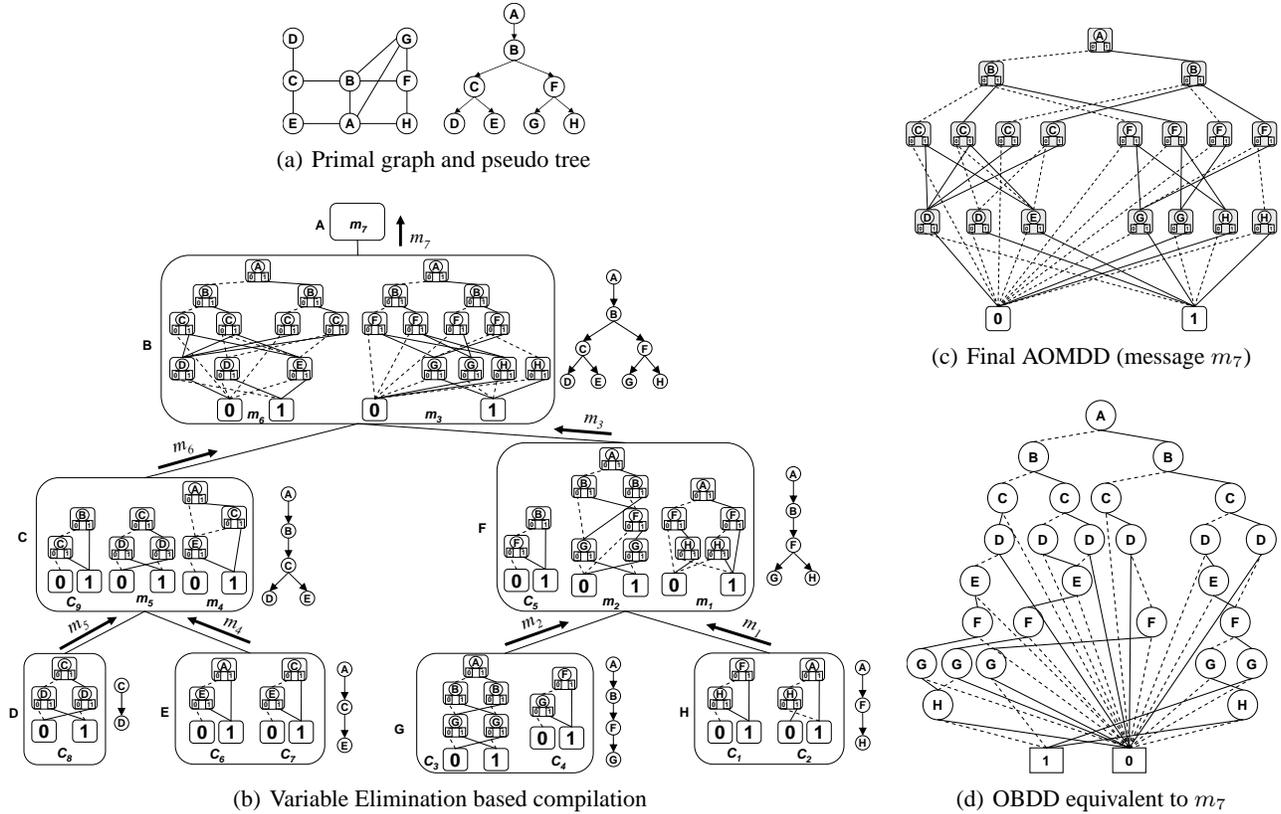

(a) Primal graph and pseudo tree

(b) Variable Elimination based compilation

(c) Final AOMDD (message $m_7$)

(d) OBDD equivalent to $m_7$

Figure 6: Execution of **VE** with AOMDDs

## 4 AND/OR Multi-Valued Decision Diagram for Constraint Networks

Constraint networks have only binary valued functions. In [13] we presented a compilation scheme for AOMDDs for constraint networks based on the Variable Elimination schedule. For completeness, we only provide below the main ideas for constraint networks, and then present the current contribution extending the AOMDD for weighted graphical models.

The context minimal graph is a data structure that is equivalent to the given graphical model, in the sense that it represents the same set of solutions, and any query on the graphical model can be answered by inspecting the context minimal graph. Our goal is to shrink the context minimal graph even further, by identifying mergeable nodes beyond those based on context. Redundant nodes can also be identified and removed.

Suppose we are given an AND/OR search graph (it could also be a tree initially). The reduction rules of OBDDs are also applicable to it, if we maintain the semantics. In particular, we have to detail the treatment of AND nodes and OR nodes. If we consider only reduction by isomorphism, then the AND/OR graph can be processed by ignoring the AND or OR attributes of the nodes. If we consider reduction by redundancy, then it is useful to group each OR node together with its AND children into a *meta-node*.

**DEFINITION 8 (meta-node)** *A nonterminal meta-node $v$ in an AND/OR search graph consists of an OR node labeled $var(v) = X_i$ and its $k_i$ AND children labeled $\langle X_i, x_{i_1} \rangle$, ..., $\langle X_i, x_{i_{k_i}} \rangle$ that correspond to its value assignments. We will sometimes abbreviate $\langle X_i, x_{i_j} \rangle$, by $x_{i_j}$. Each AND node labeled $x_{i_j}$ points to a list of child meta-nodes, $u.children_j$.*

Consider the pseudo tree in Figure 3(b). An example of meta-node corresponding to variable $A$ is given in Figure 5, assuming three values. That is just a portion of an AND/OR graph, where redundant meta-nodes were removed. For $A = 0$, the child meta-node has variable $B$. For $A = 1$, $B$ is irrelevant so the corresponding meta-node was removed, and there is an AND arc pointing to $E$ and $C$. For $A = 2$, both $B$ and $C$ are irrelevant. This example did not take into account possible weights on the OR-AND arcs.

### 4.1 Compiling AOMDDs by Variable Elimination

Consider the network defined by $\mathbf{X} = \{A, B, \ldots, H\}$, $D_A = \ldots = D_H = \{0, 1\}$ and the constraints ($\oplus$ denotes XOR): $C_1 = F \vee H, C_2 = A \vee \neg H, C_3 = A \oplus B \oplus G, C_4 = F \vee G, C_5 = B \vee F, C_6 = A \vee E, C_7 = C \vee E, C_8 = C \oplus D, C_9 = B \vee C$. The constraint graph is shown in Figure 6(a).

**Algorithm 1**: APPLY($v_1; w_1, \ldots, w_m$)

| | |
|---|---|
| **input** | : AOMDDs $f$ with nodes $v_i$ and $g$ with nodes $w_j$, based on *compatible* pseudo trees $\mathcal{T}_1$, $\mathcal{T}_2$ that can be embedded in $\mathcal{T}$. $var(v_1)$ is an ancestor of all $var(w_1), \ldots, var(w_m)$ in $\mathcal{T}$. $var(w_i)$, $var(w_j)$ are not ancestor-descendant in $\mathcal{T}$. |
| **output** | : AOMDD $v_1 \bowtie (w_1 \wedge \ldots \wedge w_m)$, based on $\mathcal{T}$. |

**1** **if** $H_1(v_1, w_1, \ldots, w_m) \neq null$ **then return** $H_1(v_1, w_1, \ldots, w_m)$
**2** **if** *(any of $v_1, w_1, \ldots, w_m$ is 0)* **then return 0**
**3** **if** *($v_1$ = 1)* **then return 1**
**4** **if** *($m$ = 0)* **then return** $v_1$
**5** create new nonterminal meta-node $u$
**6** $var(u) \leftarrow var(v_1)$ (call it $X_i$, with domain $D_i = \{x_1, \ldots, x_{k_i}\}$ )
**7** **for** $j \leftarrow 1$ **to** $k_i$ **do**
**8**     $u.children_j \leftarrow \phi$ // children of j-th AND node of $u$
**9**     **if** *( ($m$ = 1) and ($var(v_1) = var(w_1) = X_i$) )* **then**
**10**        $temp\,Children \leftarrow w_1.children_j$
**11**     **else**
**12**        $temp\,Children \leftarrow \{w_1, \ldots, w_m\}$
**13**     group nodes from $v_1.children_j \cup temp\,Children$ in several $\{v^1; w^1, \ldots, w^r\}$
**14**     **for** *each $\{v^1; w^1, \ldots, w^r\}$* **do**
**15**        $y \leftarrow$ APPLY($v^1; w^1, \ldots, w^r$)
**16**        **if** *($y$ = 0)* **then**
**17**           $u.children_j \leftarrow \mathbf{0}$; break
**18**        **else**
**19**           $u.children_j \leftarrow u.children_j \cup \{y\}$
**20**     **if** *($u.children_1 = \ldots = u.children_{k_i}$)* **then**
**21**        **return** $u.children_1$
**22**     **if** *($H_2(var(u), u.children_1, \ldots, u.children_{k_i}) \neq null$)* **then**
**23**        **return** $H_2(var(u), u.children_1, \ldots, u.children_{k_i})$
**24** Let $H_1(v_1, w_1, \ldots, w_m) = u$
**25** Let $H_2(var(u), u.children_1, \ldots, u.children_{k_i}) = u$
**26** **return** $u$

Consider the ordering $d = (A, B, C, D, E, F, G, H)$. The pseudo tree induced by $d$ is given in Fig. 6(a). Figure 6(b) shows the execution of **VE** with AOMDDs along ordering $d$. Initially, the constraints $C_1$ through $C_9$ are represented as AOMDDs and placed in the bucket of their latest variable in $d$. Each *original* constraint is represented by an AOMDD based on a chain. For bi-valued variables, they are OBDDs, for multiple-valued they are MDDs (multi-valued decision diagrams). Note that we depict metanodes: one OR node and its two AND children, that appear inside each larger square node. The dotted edge corresponds to the 0 value (the *low* edge), the solid edge to the 1 value (the *high* edge). We have some redundancy in our notation, keeping both AND value nodes and arc types (doted arcs from "0" and solid arcs from "1").

The **VE** scheduling is used to process the buckets in reverse order of $d$. A bucket is processed by *joining* all the AOMDDs inside it, using the APPLY operator (described further). However, the step of eliminating the bucket variable will be omitted because we want to generate the full AOMDD. In our example, the messages $m_1 = C_1 \bowtie C_2$ and $m_2 = C_3 \bowtie C_4$ are still based on chains, so they are still OBDDs. Note that they still contain the variables $H$ and $G$, which have not been eliminated. However, the message $m_3 = C_5 \bowtie m_1 \bowtie m_2$ is not an OBDD anymore. We can see that it follows the structure of the pseudo tree, where $F$ has two children, $G$ and $H$. Some of the nodes corresponding to $F$ have two outgoing edges for value 1.

The processing continues in the same manner The final output of the algorithm, which coincides with $m_7$, is shown in Figure 6(c). The OBDD based on the same ordering $d$ is shown in Fig. 6(d). Notice that the AOMDD has 18 nonterminal nodes and 47 edges, while the OBDD has 27 nonterminal nodes and 54 edges.

We present the APPLY algorithm for combining AOMDDs for constraints. It was shown in [13] that the complexity of the APPLY is at most quadratic in the input.

In [13] it was shown that the time and space complexity of the **VE** based compilation scheme is exponential in the treewidth of the model.

### 4.2 Compiling AOMDDs by AND/OR Search

We describe here a search based approach for compiling an AOMDD. Theorem 2 ensures that the context minimal (CM) graph can be traversed by AND/OR search in time and space $O(n \ k^{w_T^*(G)})$. When full caching is used, the trace of AND/OR search (i.e., the AND/OR graph traversed by the algorithm) is a subset of the CM graph (if pruning techniques are used, some portions of the CM graph may not be traversed). When the AND/OR search algorithm terminates, its trace is an AND/OR graph that expresses the original graphical model. We can therefore apply the reduction rules (isomorphism and redundancy) to the trace of the AND/OR search in a single bottom up pass, that has complexity linear in the size of the trace. In fact, the reduction rules can be included in the depth first AND/OR search algorithm itself: whenever the entire subgraph of a meta-node has been visited, the algorithm can check for isomorphism between the current node and metanodes of the same variable, and also check redundancy, before the search retracts to the parent meta-node. The end result will be the AOMDD of the original graphical model. The time and space complexity of this scheme is bounded in the worst case by that of exploring the CM graph, which is given in Theorem 2 (i.e., exponential in the treewidth of the model). In Section 6 we provide preliminary evaluation of the search based compilation.

## 5 AND/OR Multi-Valued Decision Diagram for Weighted Graphs

We will now describe an extension of AOMDDs to weighted graphical models, which include probabilistic graphical models. The functions defining the model can in this case take arbitrary positive real values. The AND/OR search space is well defined for such graphical models, and in particular the context minimal graph is a decision diagram that represents the same function as the model. The reduction rules (merge isomorphic nodes and reduce redundant nodes) are also well defined for weighted models (if we operate with meta-nodes), and guaranteed to produce equivalent decision diagrams. For example, isomorphic nodes should have the same variable, the same sets of children, and the same weights on their respective OR-AND arcs. If we start with the AND/OR tree and apply the isomorphism rule exhaustively, we are guaranteed to obtain a graph at least as compact as the context minimal graph. This is because OR nodes that have the same context also represent isomorphic meta-nodes when the isomorphic rule was applied exhaustively to all the levels below.

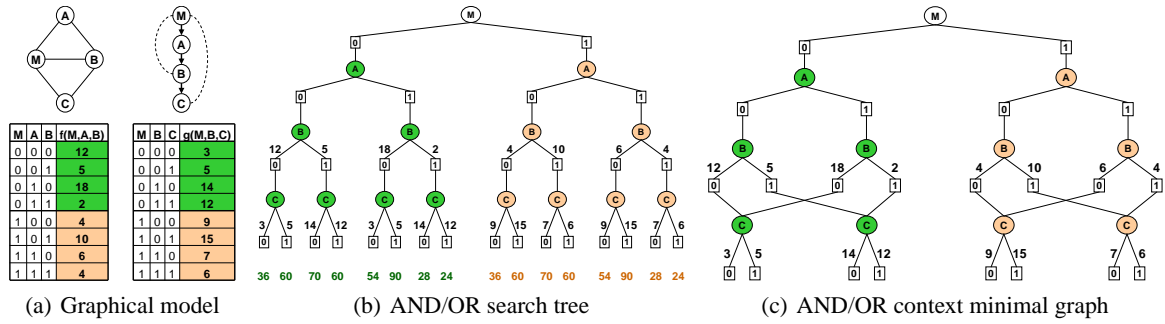

(a) Graphical model          (b) AND/OR search tree          (c) AND/OR context minimal graph

Figure 7: Weighted graphical model

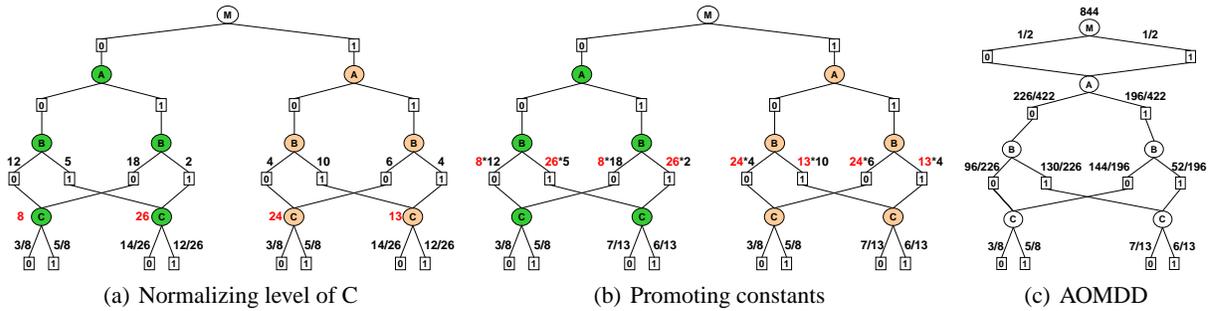

(a) Normalizing level of C          (b) Promoting constants          (c) AOMDD

Figure 8: Normalizing values bottom up

However, the property of being a canonical representation of a function is lost in the case of weighted graphs, if we only use the usual reduction rules.

Figure 7(a) shows a weighted graphical model, defined by two (cost) functions, $f(M, A, B)$ and $g(M, B, C)$. Assuming the order $(M, A, B, C)$, Figure 7(b) shows the AND/OR search tree. The arcs are labeled with function values, and the leaves show the value of the corresponding full assignment (which is the product of numbers on the arcs of the path). We can see that either value of $M$ (0 or 1) gives rise to the same function (because the leaves in the two subtrees have the same values). Therefore, the *universal model* of the conditioned subproblem for $M = 0$ is identical to that for $M = 1$. However, the two subtrees can not be identified as representing the same function by the usual reduction rules, because of different weights on the arcs. Figure 7(c) shows the context minimal graph, which has a compact representation of each subtree, but does not share any of their parts. In these figures we do not show the contours of meta-nodes, to reduce clutter.

What we would like in this case is to have a method of recognizing that the left and right subtrees corresponding to $M = 0$ and $M = 1$ represent the same function. We do this by normalizing the weights in each level, and processing bottom up by promoting the normalization constant.

In Figure 8(a) the weights on the OR-AND arcs of level $C$ have been normalized, and the normalization constant was promoted up to the OR node value. In Figure 8(b) the normalization constants are promoted upwards again by multiplication into the OR-AND weights. This process does not change the value of each full assignment, and therefore produces equivalent graphs. We can see now that some of the C level (meta) nodes are mergeable. Continuing this process gives the final AOMDD for the weighted model, in Figure 8(c).

**DEFINITION 9 (weighted AOMDD)** *A weighted AOMDD is an AND/OR graph (with meta-nodes), where for each OR node, the emanating OR-AND arcs have an associated weight, such that their sum is 1, and the root meta-node has a weight (the resulting normalization constant). The terminal nodes are just* **0** *and* **1**.

The following theorem ensures the that the weighted AOMDD is a canonical representation.

**THEOREM 3** *Given two equivalent weighted graphical*

| Network | Class | (n, d) | e | (w*, h) | Zeros (%) | Time (sec) | #cm | #aomdd | Ratio |
|---------|-------|--------|---|---------|-----------|------------|-----|--------|-------|
| cpcs54 | | (54, 2) | 5 | (12, 20) | 0.00 | 1.02 | 24,422 | 23,845 | 1.02 |
| cpcs179 | | (179, 4) | 5 | (7, 13) | 0.32 | 110.24 | 31,389 | 9,702 | **3.24** |
| cpcs360b | CPCS | (360, 2) | 50 | (16, 21) | 0.68 | 105.49 | 159,863 | 155,670 | 1.03 |
| cpcs422b | | (422, 2) | 15 | (16, 33) | 0.78 | 57.74 | 429,270 | 25,754 | **16.67** |
| c432 | | (432, 2) | 40 | (21, 37) | 48.98 | 63.38 | 345,401 | 119,109 | **2.90** |
| c499 | ISCAS | (499, 2) | 40 | (20, 50) | 47.76 | 20.45 | 106,659 | 105,921 | 1.01 |
| s386 | | (172, 2) | 10 | (17, 30) | 48.18 | 6.27 | 33,806 | 4,425 | **7.64** |
| s953 | | (440, 2) | 120 | (29, 51) | 45.45 | 180.60 | 183,199 | 46,890 | **3.91** |
| 90-10-1 | | (100, 2) | 10 | (10, 31) | 42.42 | 0.78 | 6,812 | 3,544 | 1.92 |
| 90-14-1 | GRID | (196, 2) | 10 | (16, 60) | 45.25 | 68.16 | 317,527 | 73,245 | **4.34** |
| 90-16-1 | | (256, 2) | 20 | (18, 66) | 43.99 | 278.05 | 1,305,719 | 256,990 | **5.08** |
| 90-24-1 | | (576, 2) | 120 | (17, 77) | 44.66 | 102.78 | 569,282 | 179,668 | **3.17** |
| EA3 | | (711, 5) | 0 | (14, 46) | 37.49 | 4.55 | 29,144 | 15,654 | 1.86 |
| EA4 | | (775, 5) | 0 | (16, 62) | 37.34 | 5.88 | 21,246 | 14,653 | 1.45 |
| EA5 | LINKAGE | (944, 5) | 0 | (13, 53) | 36.46 | 6.75 | 25,973 | 12,377 | **2.10** |
| EA6 | | (1136, 5) | 0 | (17, 79) | 36.60 | 18.12 | 43,202 | 27,960 | 1.55 |
| bm-05-01 | | (700, 2) | 100 | (17, 36) | 49.17 | 105.19 | 275,445 | 115,327 | **2.39** |
| bm-05-02 | | (1418, 2) | 100 | (16, 50) | 49.44 | 60.97 | 143,866 | 74,818 | 1.92 |
| mm-03-08-03 | PRIMULA | (1220, 2) | 100 | (17, 53) | 48.94 | 748.71 | 898,631 | 384,049 | **2.34** |
| mm-04-08-03 | | (1418, 2) | 300 | (17, 49) | 47.89 | 48.44 | 158,584 | 50,718 | **3.13** |

Table 1: Results for experiments with 20 belief networks from 5 problem classes.

*models that accept a common pseudo tree $\mathcal{T}$, normalizing arc values together with exhaustive application of reduction rules yields the same AND/OR graphs.*

The proof is omitted here for space reasons. We only mention that the proof is by structural induction bottom up over the layers of the AND/OR graph.

The APPLY algorithm needs minimal modifications now to operate on weighted AOMDDs. The hash function $H_2$, which hashes meta-nodes, has to take as extra arguments the weights of the meta-node. Similarly, when checking redundancy in line 21, the weights should also be equal for the node to be redundant, and their common value has to be promoted by multiplication. When checking isomorphism in line 23, the corresponding weights can be checked via the hash function $H_2$. The same **VE** schedule can now be used to compile an AOMDD for a weighted graphical model.

## 6   Experimental Evaluation

Our experimental evaluation is in preliminary stages, but the results we have are already encouraging. We ran the search based compile algorithm, by recording the trace of the AND/OR search, and then reducing the resulting AND/OR graph bottom up. In these results we only applied the reduction by isomorphism and still kept the redundant meta-nodes.

Table 1 shows the results for 20 belief networks from 5 problem classes: medical diagnosis (CPCS), digital circuits (ISCAS), deterministic grid networks (GRID), genetic linkage analysis (LINKAGE) as well as relational belief networks (PRIMULA). For each network we chose randomly $e$ variables and set their values as evidence. For each query we recorded the compilation time in seconds, the number of OR nodes in the context minimal graph explored (#cm) and the size of the resulting AOMDD (#aomdd). In addition, we also computed the compression ratio of the AOMDD structure as $ratio = \#cm/\#aomdd$. We also report the number of variables (n), domain size (d), induced width ($w^*$), pseudo tree depth (h), as well as the percentage of zero probability tuples (zeros (%)) for each test instance.

We see that in a few cases the compression ratio is significant (e.g., cpcs422b 16.67%, s386 7.64%). Our future work will include the reduction rule by redundancy, as well as the compilation algorithm by Variable Elimination schedule.

## 7   Conclusion and Discussion

We presented the new data structure of weighted AOMDD, as a target for compilation of weighted graphical models. It is based on AND/OR search spaces and Binary Decision Diagrams. We argue that the AOMDD has an intuitive structure, and can easily be incorporated into other already existing algorithm (e.g., join tree clustering). We provide two compilation methods, one based on Variable Elimination and the other based on search, both being time and space exponential in the treewidth of the graphical model. The preliminary experimental evaluation is quite encouraging, and shows the potential of the new AOMDD data structure.

Compiling graphical models into weighted AOMDDs also extends decision diagrams for the computation of semiring valuations [18], from linear variable ordering into tree-based partial ordering. This provides an improvement of the complexity guarantees to exponential in treewidth, rather than pathwidth.

There are various lines of related research. We only mention here: deterministic decomposable negation normal form (d-DNNF) [7]; case factor diagrams [14]; compilation of CSPs into tree-driven automata [10]; and the recent work on compilation [16, 4]. We think that our framework using AND/OR search graphs has a unifying quality that helps make connections among seemingly different compilation techniques.

The approach of compiling graphical models into AOMDDs may seem to go against the current trend in model checking, which moves away from BDD-based algorithms into CSP/SAT based approaches. However, algorithms that are search-based and compiled data-structures such as BDDs differ primarily by their choices of time vs memory. When we move from regular OR search space to an AND/OR search space the spectrum of algorithms available is improved for all time vs memory decisions. We believe that the AND/OR search space clarifies the available choices and helps guide the user into making an informed selection of the algorithm that would fit best the particular query asked, the specific input function and the computational resources.

## Acknowledgments

We thank Radu Marinescu for his help with the experimental evaluation. This research was supported in part by the NSF grant IIS-0412854.